\documentclass[letterpaper, 10 pt, conference]{ieeeconf}  % Comment this line out if you need a4paper

\IEEEoverridecommandlockouts                              % This command is only needed if 
\usepackage{graphics} % for pdf, bitmapped graphics files
\usepackage{epsfig} % for postscript graphics files
\usepackage{mathptmx} % assumes new font selection scheme installed
\usepackage{times} % assumes new font selection scheme installed
\usepackage{amsmath, dsfont} % assumes amsmath package installed
\usepackage{amssymb}  % assumes amsmath package installed
\usepackage{subcaption}
\usepackage{textcomp}
\usepackage{stfloats}
\usepackage{url}
\usepackage{verbatim}
\usepackage{graphicx}
\usepackage{hyperref}
\usepackage{bbm}
\usepackage{bm}
\usepackage{color}
\usepackage{algorithm}
\usepackage{algpseudocode}
\algnewcommand\algorithmicinput{\textbf{\quad Input:}}
\algnewcommand\INPUT{\item[\algorithmicinput]}
\algnewcommand\algorithmicfind{\qquad Find}
\algnewcommand\FIND{\item[\algorithmicfind]}

\newcommand{\mR}{{\mathbb R}}

\newcommand{\bU}{{\mathbf U}}

\newcommand{\bD}{{\mathbf D}}
\newcommand{\bQ}{{\mathbf Q}}

\newcommand{\bh}{{\mathbf h}}
\newcommand{\br}{{\mathbf r}}
\newcommand{\bs}{{\mathbf s}}
\newcommand{\bq}{{\mathbf q}}
\newcommand{\bv}{{\mathbf v}}

\newcommand{\bk}{{\mathbf k}}

\newcommand{\bK}{{\mathbf K}}
\newcommand{\bp}{{\mathbf p}}

\newcommand{\bc}{{\mathbf c}}

\newcommand{\bff}{{\mathbf f}}

\newcommand{\bx}{{\mathbf x}}

\newcommand{\bX}{{\mathbf X}}

\newcommand{\bg}{{\mathbf g}}
\newcommand{\bA}{{\mathbf A}}

\newcommand{\bu}{{\mathbf u}}

\newcommand{\bJ}{{\mathbf J}}
\newcommand{\btheta}{{\boldsymbol \theta}}

\newcommand{\btau}{{\boldsymbol \tau}}

\newtheorem{theorem}{Theorem}
\newtheorem{definition}{Definition}

\newtheorem{remark}{Remark}

\newtheorem{problem}{Problem}

\title{\LARGE \bf
Safe Motion Planning for Quadruped Robots Using Density Functions
}

\author{Sriram S.K.S Narayanan$^{*1}$, Andrew Zheng$^{*1}$ and Umesh Vaidya$^{1}$ % <-this % stops a space
\thanks{*These authors contributed equally}% <-this % stops a space
\thanks{Financial support from of NSF CPS award 1932458 and NSF
2031573 is greatly acknowledged. Sriram S.K.S Narayanan, Andrew Zheng, and Umesh Vaidya are with the Department of Mechanical Engineering, Clemson University, Clemson, SC 29630, USA
        {\tt\small email: sriramk@clemson.edu; azheng@clemson.edu; uvaidya@clemson.edu}}%
}

\begin{document}
\maketitle
\thispagestyle{empty}
\pagestyle{empty}

%%%%%%%%%%%%%%%%%%%%%%%%%%%%%%%%%%%%%%%%%%%%%%%%%%%%%%%%%%%%%%%%%%%%%%%%%%%%%%%%
\begin{abstract}
This paper presents a motion planning algorithm for quadruped locomotion based on density functions. We decompose the locomotion problem into a high-level density planner and a model predictive controller (MPC). Due to density functions having a physical interpretation through the notion of occupancy, it is intuitive to represent the environment with safety constraints. Hence, there is an ease of use to constructing the planning problem with density. The proposed method uses a simplified the model of the robot into an integrator system, where the high-level plan is in a feedback form formulated through an analytically constructed density function. We then use the MPC to optimize the reference trajectory, in which a low-level PID controller is used to obtain the torque level control. The overall framework is implemented in simulation, demonstrating our feedback density planner for legged locomotion. The implementation of work is available at \url{https://github.com/AndrewZheng-1011/legged_planner}

\end{abstract}

%%%%%%%%%%%%%%%%%%%%%%%%%%%%%%%%%%%%%%%%%%%%%%%%%%%%%%%%%%%%%%%%%%%%%%%%%%%%%%%%
\section{Introduction}\label{sec:intro}
Quadruped research has seen considerable growth over the past few years \cite{wensing2022optimization}. This is highlighted through the DARPA Subterranean Challenge, where top contenders used legged robots as a core component to their underground autonomy challenge \cite{tranzatto2022cerberus}, \cite{chung2022into}. This shift in legged systems rather than mobile bases can be attributed to advancements in the algorithmic development and improvements in real-time optimization that have improved the legged systems' capabilities in traversing over challenging environments. {\color{black} However, despite the successes, navigating under safety constraints for legged robots in complex environments still has its difficulties.}

{\color{black}
Sample-based planners such as rapidly exploring random tree search (RRT) and probabilistic roadmaps (PRM) \cite{lavalle1998rapidly, amato1996randomized} are very efficient in generating safe motion plans. Here, safety is enforced through local collision checks at each sample point, which can be inefficient for complex environments. More recently control barrier functions (CBF) have seen great success in enforcing safety. The barrier functions act as a safety certificate and can be implemented as a safety filter using a quadratic program (QP) \cite{ames2019control,ames2016control}. However, designing these CBFs for high-dimensional nonlinear systems are nontrivial \cite{molnar2021model, singletary2022onboard}. 

Alternatively, the motion planning problem can be formulated in the dual space of density. It has been shown that density functions can be viewed as an alternative safety certificate, enabling synthesis of safe controllers \cite{rantzer2004analysis,dimarogonas2007application,loizou2008density}. Moreso, a convex formulation for synthesizing safe controllers using navigation measure was introduced in \cite{vaidya2018optimal}. These works were extended to the data-driven setting based on the linear transfer operators with safety constraints in \cite{yu2022data,10081458}. However, recently, an {\color{black}analytically} constructed density functions was proposed that jointly solves the obstacle avoidance and convergence problem through a feedback control fashion in \cite{zheng2023safe}.
}

% In general, the quadruped locomotion problem is decomposed into a high-level global planner, which provides collision-free paths for the floating base, and a low-level controller for designing the required joint torques for each leg to track the reference trajectory. 
% This work uses the density functions proposed in \cite{zheng2023safe} as a tool to design safe high-level trajectories. 

\begin{figure}
     \centering
     \includegraphics[width = \linewidth]{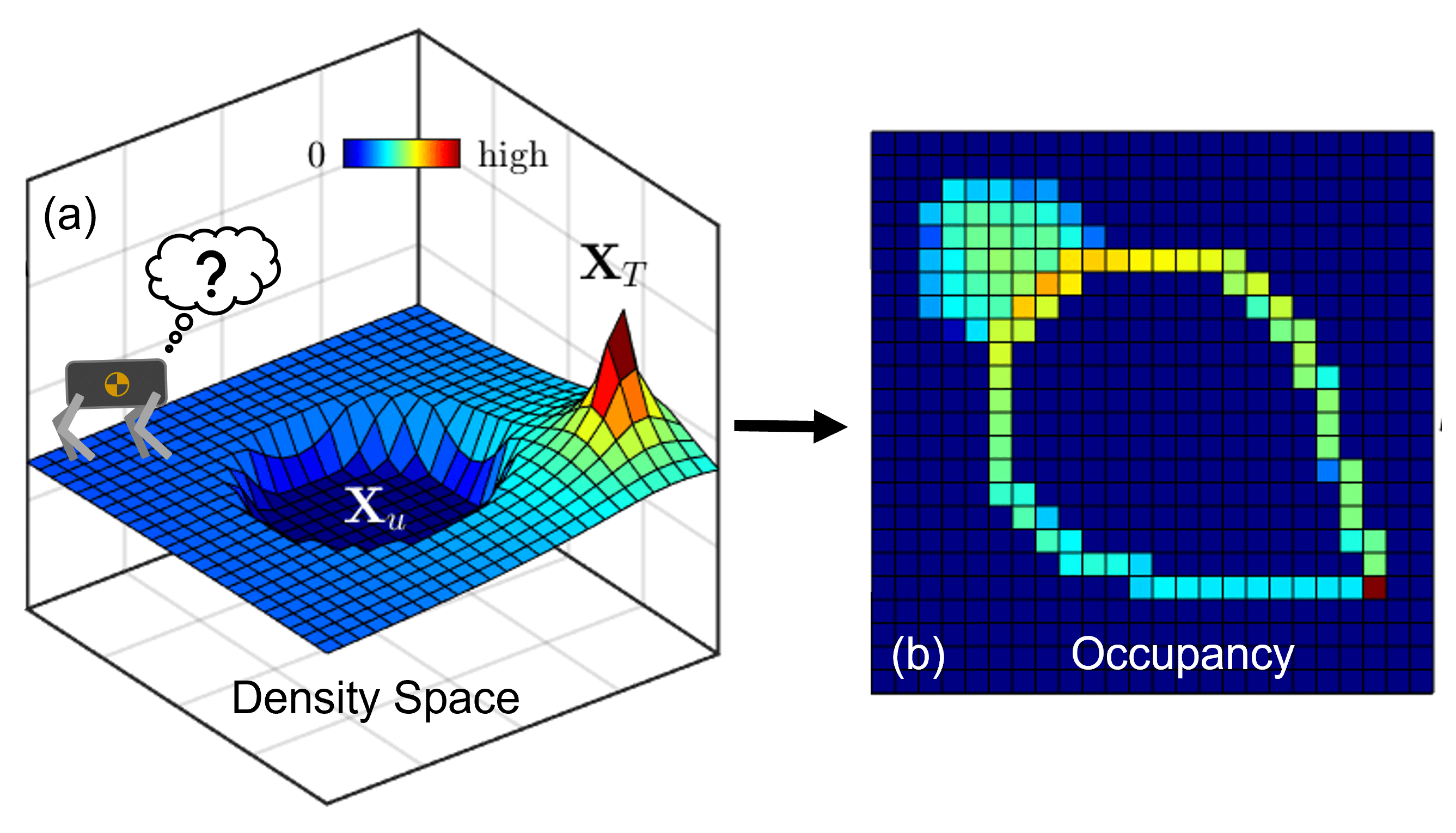}
     \caption{Motion Planning framework using density where (a) shows the density function defined on the environment, and (b) shows occupancy measure, which physically denotes the duration of system trajectories occupying the set.}
     \label{fig:dens_nav_diagram}
 \end{figure}

{\color{black} The main contribution of this work is to design a safe motion plan for quadruped robots using density functions. Unlike potential functions, which are known to exist but hard to construct for arbitrary environments, \cite{zheng2023safe} provides a systematic way to construct density functions for any arbitrary environment.} As the density function formulated has nice physical interpretation with occupation as seen in Figure \ref{fig:dens_nav_diagram} (i.e. the proposed density function formulated has zero density at the unsafe set, hence zero occupation in the set), we exploit this occupancy-based physical interpretation of the density function to design collision-free plans. More specifically, the plan is obtained as the gradient of the density function and can be received in a feedback form. From the received feedback plan, we then use a nonlinear model predictive controller (NMPC) as {\color{black}an} optimizer to obtain reference foot forces to track the high-level reference trajectories. Finally, the optimized foot forces are converted to joint torques using a low-level PID controller.

The rest of the paper is organized as follows. In Section \ref{sec:prelims}, {\color{black} we introduce the preliminaries of formulating the almost everywhere navigation problem. In section \ref{sec:construction_nav_dens}, we provide the analytical construction of density functions}. In Section \ref{sec:integrator}, we provide simulation results of the proposed density-based planner {\color{black}for} a single integrator system. In Section \ref{sec:quad}, we integrate the density-based planner with the quadruped locomotion framework using ROS and Gazebo. Finally, in Section \ref{sec:conclusions}, we conclude this work with some limitations and future directions.

%%%%%%%%%%%%%%%%%%%%%%%%%%%%%%%%%%%%%%%%%%%%%%%%%%%%%%%%%%%%%%%%%%%%%%%%%%%%%%%%
\section{Preliminaries}\label{sec:prelims}
\noindent {\bf Notations}: The following notations will be used in this paper. $\mathbb{R}^n$ denotes the $n$ dimensional Euclidean space, $\bx \in \mathbb{R}^n$ denotes a vector of system states, $\bu \in \mathbb{R}^n$ is a vector of control inputs. Let $\bX \subset \mR^n$ be a bounded subset that denotes the workspace for the robot. $\bX_0, \; \bX_T,\; \bX_{u_k} \subset \bX$,  for $k=1,\ldots, L$  denote the initial, target, and unsafe sets, respectively. With no loss of generality, we will assume that the target set is a single point set located at the origin, i.e., $\bX_T=\{0\}$. $\bX_u=\cup_{k=1}^L\bX_{u_k}$ defines the unsafe set and $\bX_{s}:=\bX\setminus \bX_u$ defines the safe set. We use $\mathcal{C}^k(\bX)$ to denote the space of all $k$-times differentiable functions of $\bx$. We use ${\cal M}(\bX)$ to denote the space of all measures on $\bX$ and $m(\cdot)$ to denote the Lebesgue measure. $\mathds{1}_A(\bx)$ denotes the indicator function for set $A\subset \bX$. 

%%%%%%%%%%%%%%%% Problem statement %%%%%%%%%%%%%%%%%%%%%%%%%%%
Next, we formally define the motion planning problem used in the rest of this paper.
\begin{problem}[\textrm{{\color{black}\cite{zheng2023safe}}}] (Almost everywhere (a.e.) navigation problem)\label{problem1}
The objective of this problem is to design a smooth feedback control input $\bu=\bk(\bx)$ to drive the trajectories of the dynamical system
 \begin{align}\label{sys}
 \dot \bx=\bu,\;\;\;\;\;%\bu=\bk(\bx)
 \end{align}
from almost every initial condition (w.r.t. Lebesgue measure)  from the initial set $\bX_0$ to the target set $\bX_T$ while avoiding the unsafe set $\bX_u$. 
\end{problem}
% \begin{assumption} We assume that there exists a feedback controller that solves the a.e. navigation problem as stated above.
%     % A feedback control solution for the dynamical system exists for almost every initial condition that drives trajectories to $\bX_T$. 
% \label{assum:feasibility}
% \end{assumption}

Density functions are a physically intuitive way to solve the a.e. navigation problem presented in Problem \ref{problem1}. Specifically, they can be used to represent arbitrary unsafe sets and generate safe trajectories for the system \eqref{sys}. In this paper, we define safe trajectories as the ones that have zero occupancy in the unsafe set $\bX_u$. The formal definition of occupancy used in this paper is defined below.
\begin{definition}[{\color{black}\cite{vaidya2018optimal}}] (Occupancy of a set) Let $\bA\subset \bX$ be a measurable set. The occupancy of the system \eqref{sys} trajectories in the set $\bA$ while traversing from the initial set $\bX_0$ to the target set $\bX_T$ is defined as 
\begin{align}
\mu_\bA:=\int_0^\infty \int_{\bX} {\mathds 1}_\bA(\bs_t(\bx))\mathds{1}_{\bX_0}(\bx)dt
\end{align}
{\color{black} where $\bs_t(\bx)$ is the solution to system defined in \eqref{sys}}
\end{definition}
\vspace{2mm}

%%%%%%%%%%%%%%%%%%%%%%%%%% Density Function %%%%%%%%%%%%%%%%%%%%%%%
\section{Construction of Density Functions for Safe Motion Planning}\label{sec:construction_nav_dens}
As introduced in \cite{vaidya2018optimal}, the navigation measure has a physical interpretation of occupancy, where the measure of any set is equal to the occupancy of the system trajectories in the set as shown in Figure \ref{fig:dens_nav_diagram}. Hence, zero occupancy in a set implies system trajectories not occupying that particular set. So by ensuring that the navigation measure is zero on the obstacle set and maximum on the target set, it is possible to induce dynamics whereby the system trajectories will reach the desired target set while avoiding the obstacle set. We exploit this occupancy-based interpretation in the construction of analytical density functions.

Given a valid density function for the system \eqref{sys}, a safe trajectory can be obtained based on Rantzer's dual Lyapunov theorem stated below \cite{rantzer2001dual}.

\begin{theorem} \label{thm:conv_and_avoidance} Consider a dynamic system defined by $\dot{\bx} = \bff(\bx)$ with the vector field $\bff\in \mathcal{C}^1(\mathbb{R}^n, \mathbb{R}^n)$ and $\bff(\bx^{\star})= \mathbf{0}$ be an equilibrium point (or target set). Suppose there exists a non-negative function $\rho(\bx) \in \mathcal{C}^1(\mathbb{R}^n \setminus \bX_T, \mathbb{R})$ such that $\frac{\rho(\bx)\bff(\bx)}{||\bx-\bx^\star||}$ is integrable on $\{\bx \in \bX: ||\bx-\bx^\star||\geq 1\}$ and
\begin{equation}\label{eq:thm1}
    [\nabla \cdot (\bff\rho)](\bx) >  0 \;\;\;a.e.\;\; \bx\in \bX\setminus \bX_0
\end{equation}
\begin{align}\label{eq:thm1_pt2}
\int_{\bX_0}\rho(\bx)d\bx=0
\end{align}
then, for almost all initial states $\bX_0$, the solution {\color{black}$s_t(\bx)$}, converges to $\bx^\star$ as $t \to \infty$ and 
\[\mathds{1}_{\bX_0}(\bx(t))=0\]
\end{theorem}

% In this paper, we introduce a novel construction of density functions which includes information of the unsafe sets $\bX_u$ and guarantees collision-free trajectories for the system in \eqref{sys} with the controller as the positive gradient of the proposed density function, $\bu(\bx)=\nabla\rho(\bx)$.
 
Although density functions can represent any arbitrarily shaped obstacles, we restrict the focus of this paper to circular obstacles. For each obstacle $k$, we start with constructing the unsafe set $\bX_{u_k}$, where the boundary of the unsafe set is described in terms of the zero-level set of a function. For a circular obstacle {\color{black} of radius $r_k$,} the unsafe set $\bX_{u_k}$ is defined as follows
\begin{align}
\bX_{u_k}=\{\bx\in \bX: \|\bx-\bc_k\|\leq r_k\} \nonumber
\end{align}

Next, for each obstacle, we define a circular sensing region $\bX_{s_k}$ with radius $s_k$ that encloses the unsafe set $\bX_{u_k}$. Inside this region, the robot starts to react to the unsafe set.
\begin{align}
\bX_{s_k}=\{\bx\in \bX: \|\bx-\bc_k\|\leq s_k\} \setminus \bX_{u_k} \nonumber 
\end{align}
Now we can define a positive scalar-valued density function $\rho(\bx)$, which takes the following form
\begin{align}
\rho(\bx)=\frac{\prod_{k=1}^L \Phi_k(\bx)}{V(\bx)^\alpha}\label{density_fun}
\end{align}
Here, the function $V(\bx)$ is the distance function that measures the distance from state $\bx$ to the target set, i.e. the origin, and $\alpha$ is a positive scalar. In this paper, we assume $V(\bx)$ to be of the form $V(\bx)=\|\bx\|^2$. The inverse bump function $\Phi_k(\bx)$ is a smooth $\mathcal{C}^\infty$ function that captures the geometry of the unsafe set $\bX_{u_k}$ and can be constructed using the following sequence of functions. We first define an elementary $\mathcal{C}^\infty$ function $f$ as follows \cite{tu2011manifolds}
\begin{align} \label{eq:elementary_f}
    &f(\tau) = \begin{cases}
        \exp{(\frac{-1}{\tau})}, &\tau > 0 \nonumber \\
        0, & \tau \leq 0
    \end{cases}
\end{align}
where $\tau \in \mathbb{R}$. Next, we construct a smooth version of a step function $\bar{f}$ from $f$ as follows
\begin{equation} \label{eqn:smooth_step_g}
    \bar{f}(\tau) = \frac{f(\tau)}{f(\tau)+f(1-\tau)} \nonumber
\end{equation}
To incorporate the geometry of the environment, we define a change of variables such that $\phi_k(\bx) = \bar{f}\Bigl(\frac{\|\bx-\bc_k\|^2-r_k^2}{s_k^2-r_k^2}\Bigr)$. The resulting function $\Phi_k(\bx)$ take the following form,

\begin{align}
\Phi_k(\bx) = \begin{cases}
    0, & \bx \in \bX_{u_k}  \\
    \phi_k(\bx), & \bx \in \bX_{s_k}\\
    1, & \rm{otherwise}
\end{cases} \nonumber
\end{align} 

Note that $\alpha$ and $\bs_k$ are scalar tuning parameters that can be used to obtain trajectories with the desired behavior.

%%%%%%%%%%%%%%%%%%%%%%%%%%%%%%%%%%%%%%%%%%%%%%%%%%%%%%%%%%%%%%%%%%%%%%%%%%%%%%%%
\section{Safe Motion Planning for a Holonomic Single Integrator} \label{sec:integrator}
Given the construction of $\rho(\mathbf{x})$ in (\ref{density_fun}), we design a controller for navigation as the positive gradient of the density function $\rho(\bx)$, i.e.,
\begin{align}
\dot \bx= \nabla \rho(\mathbf{x})  \label{eq:control}
\end{align}
The controller defined in equation \eqref{eq:control} will converge to the goal if the density function $\rho(\bx)$ satisfies almost everywhere navigation properties \cite{rantzer2001dual}. The density function proposed in equation \eqref{density_fun} can be used for obstacle avoidance while also satisfying the convergence properties. {\color{black} See \cite{zheng2023safe} for detailed proof.}

\begin{remark} \label{remark:local_controller}
 {\color{black} We make the following modification to 
(\ref{eq:control}) to ensure that the vector field is well-defined and the origin is locally asymptotically stable in ${\cal B}_\delta$.    $\dot \bx = \left[1-\Bar{f}(\tau)\right]\nabla \rho(\bx) - \Bar{f}(\tau)\bx$ where, $\bar f$ is as defined in (\ref{eqn:smooth_step_g}).} {\color{black} With this modification, we will continue to work with (\ref{eq:control}) with the assumption that the origin is locally asymptotically stable in ${\cal B}_\delta$ for (\ref{eq:control}). }
\end{remark}

\begin{figure}
    \centering
    \includegraphics[width = \linewidth]{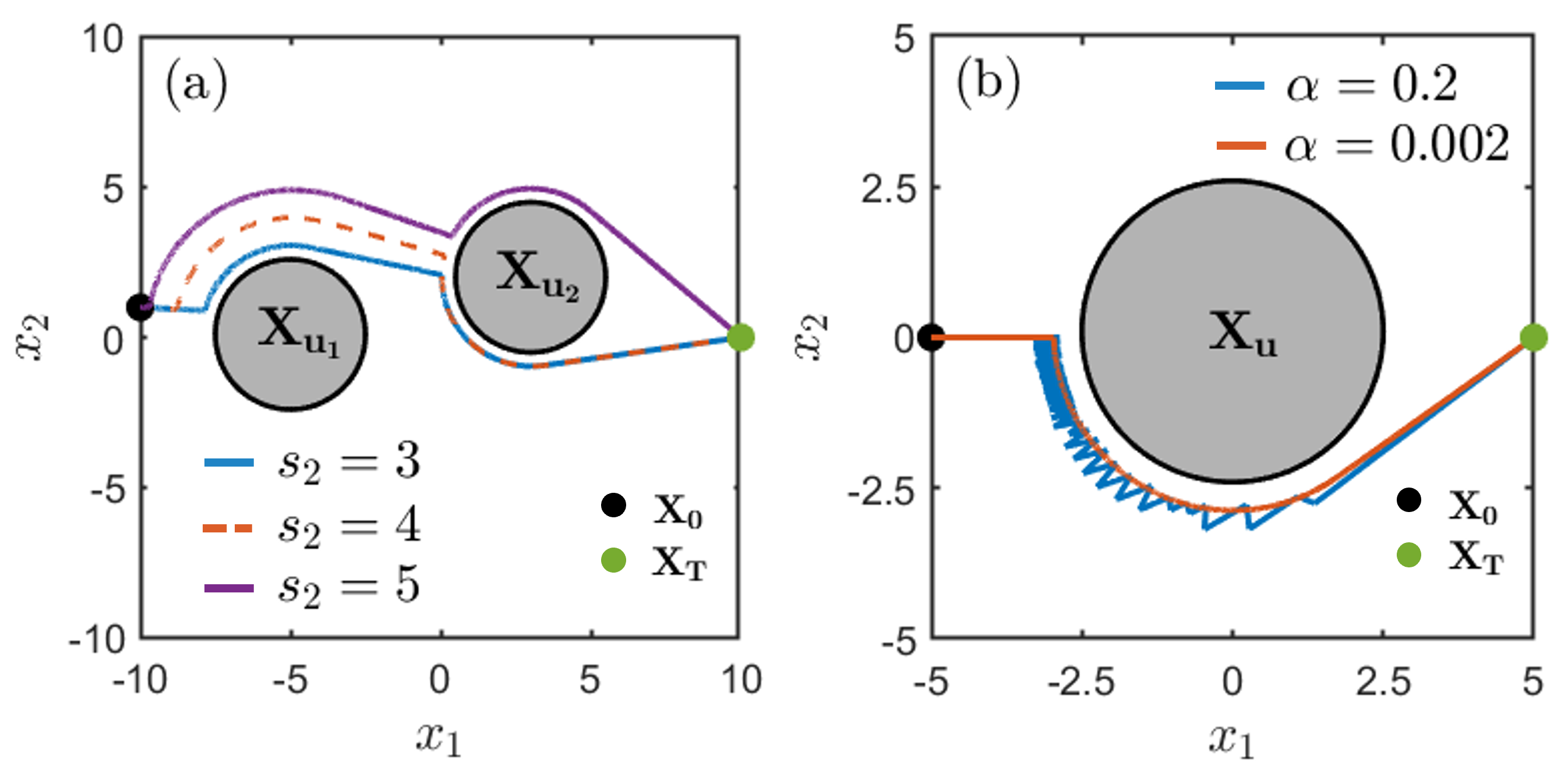}
    \caption{Solution trajectories for varying values of (a) sensing radius $s_2$ and (b) tuning parameter $\alpha$. }
    \label{fig:saddle_points}
\end{figure}

 We demonstrate this using a simple example as shown in Figure \ref{fig:saddle_points}. The environment is defined with the target set at {\color{black}$\bX_T = (10,0)$} and a circular unsafe set $\bX_u$ and circular sensing region $\bX_{s}$.

Figure \ref{fig:saddle_points} illustrates the convergence properties of the proposed controller for various tuning parameters. In Figure \ref{fig:saddle_points}a, the environment is set up with circular unsafe sets $\bX_{u_1}, \; \bX_{u_2}$ with radius $r_1=2.5$ and $r_2=2.5$ respectively (with $\alpha=0.2$ and simulation step size $dt=0.01$ s{\color{black})}. The circular sensing region for the $\bX_{u_1}$ has a radius of $s_1=3$ while for $\bX_{u_2}$, it is varied between $s_2=[3,\;4,\;5]$. It can be seen that different values of $s_2$ can generate different solution trajectories while avoiding unsafe sets. In Figure \ref{fig:saddle_points}b, the tuning parameter $\alpha$ is varied between $\alpha=0.2$ and $\alpha=0.002$ (for a simulation step size $dt=0.1$ s). It can be noted that $\alpha=0.002$ generates a smoother trajectory for the corresponding step size.

% \begin{table}[ht]
% \caption{Density Parameters}
% \centering
% \begin{tabular}{ c c c c } 
%  \hline
%  Parameter & Value \\
% \hline
%  $\alpha$      & 0.2 \\
%  $r$           & 1 \\
%  $s$           & 2 \\
%  Control Gain       & 25 \\
% \hline
% \end{tabular}
% \label{tab:gains}
% \end{table}

%%%%%%%%%%%%%%%%%%%%%%%%%%%%%%%%%%%%%%%%%%%%%%%%%%%%%%%%%%%%%%%%%%%%%%%%%%%%%%%%
%%%%%%%%%%%%%%%%%%%%%%%%%%%%%%%%%%%%%%%%%%%%%%%%%%%%%%%%%%%%%%%%%%%%%%%%%%%%%%%%
\section{Safe Motion Planning for Quadrupeds} \label{sec:quad}
Quadruped locomotion can be described as a hybrid system switching between swing and stance phase dynamics. The switching logic is determined by a Finite State Machine (FSM) based on a contact detection algorithm. This system is under-actuated since there is no direct actuation along the direction of motion. The robot must exert ground reaction forces (GRFs) at each foot in contact determined by a model predictive controller (MPC) to propel its base forward to follow a reference trajectory given by the global planner. Finally, a low-level controller is used to generate the required joint torques to track the GRFs. 
\begin{figure}
    \centering
  \includegraphics[width=1\linewidth]{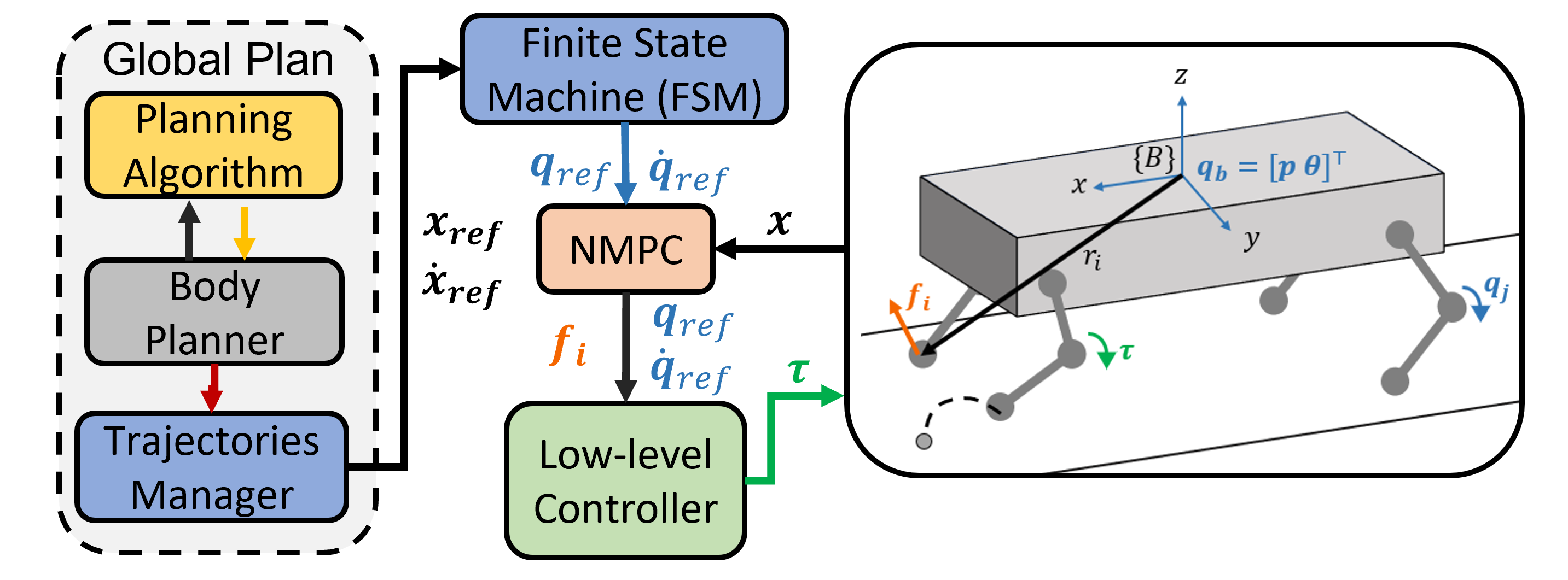}
  \caption{Hierarchical planning and control structure for quadruped locomotion.}
  \label{fig:quad_diagram}
\end{figure}
\subsection{Global Planner Architecture}
This section discusses the planning architecture to map the proposed density-based planner for holonomic systems to quadruped locomotion. {\color{black}The main contribution of this work is the integration of a density-based safe motion-planning architecture based on \cite{zheng2023safe} into the OCS2 framework \cite{OCS2}, as seen in Figure \ref{fig:quad_diagram}}. The goal of the motion planning architecture is to create an interface that extends the density-based motion plan for holonomic systems to a high-level global plan suitable for quadruped locomotion. The architecture relies mainly on the body planner to handle the logic of receiving plans, mapping the plan to a centroidal model, and sending the lifted plan to a trajectory manager to handle trajectories as a reference to the FSM and MPC. Details of the implementation is available on GitHub\footnote{\href{https://github.com/AndrewZheng-1011/legged_planner}{{https://github.com/AndrewZheng-1011/legged\_planner}}}.

%%%%%%%%%%%%%%%%%%%%%%%%%% Quadruped Locomotion %%%%%%%%%%%%%%%%%%%%%%%
\subsection{Model Predictive Control}
In this section, we introduce centroidal dynamics which is used to model the movement of a rigid body, and a nonlinear model predictive controller (NMPC) which is used to generate (GRFs) required to track a collision-free center of mass reference trajectory. Finally, we also present a quadratic programming (QP) based low-level controller to generate the joint torques required to track the desired GRFs.

The centroidal model describes the evolution of the floating base as a result of the net change of momentum of the full-order system. The state and input vectors are defined by 
\begin{align} \label{eq:mapping}
    &\bx := [\bh_{com} \hspace{1mm} \bq_b \hspace{1mm} \bq_j]^\top \in \mathbb{R}^{12+n} \nonumber \\ 
    &\bu := [\bff_{c_i} \hspace{1mm} \bv_j] \in \mathbb{R}^{3n_c + n_j} \nonumber
\end{align}
where $\bh_{com} = [\bh_{lin} \hspace{1mm} \bh_{ang}] \in \mathbb{R}^6$ is the centroidal momentum and $\bh_{lin}$, $\bh_{ang}$ are the linear and angular momentum respectively. The states $\bq_b = [\bp \hspace{1mm} \btheta] \in \mathbb{R}^6$ represents the position and orientation of the floating base. The states $\bm{q}_j \in \mathbb{R}^n$ represent the joint angles for each foot. Each leg $i$ in contact generates a ground reaction force $\bff_{c_i} \in \mathbb{R}^3$, and $v_j$ is the angular velocity of each joint. Here, $n_c$ and {\color{black}$n_j$} are the number of contact points and the number of leg joints, respectively. The centroidal dynamics are given by 
\begin{align}
    \dot{\bh}_{com} = \begin{bmatrix}
        &\sum_{i=1}^{n_c} \bff_{c_i} + m\bg \\
        &\sum_{i=1}^{n_c} \br_{com} \times \bff_{c_i} + \btau_{c_i}
    \end{bmatrix} \nonumber
\end{align}
where $m$ is the mass of the robot, $\bg = [0 \hspace{2mm} 0 \hspace{2mm} g]^\top$ denotes the gravity vector. $\bff_{c_i}$ and  $\btau_{c_i}$ represent the contact forces and torques applied by the environment on the floating base. Further, we use the centroidal momentum matrix $\bm{A}(\bm{q}) = [\bA_b(\bq) \hspace{1mm}  \bA_j(\bq)] \in \mathbb{R}^{6 \times (6+{\color{black}n_j})}$ as introduced in \cite{orin2013centroidal} such that
\begin{align}
    \bh_{com} = \begin{bmatrix} \bA_b(\bq) & \bA_j(\bq)
    \end{bmatrix} \begin{bmatrix}
        \dot{\bq}_b \\ \dot{\bq}_j
    \end{bmatrix} \nonumber
\end{align}
Finally, the dynamics of floating base $\bq_b$ is augmented with the kinematics of the joints $\bq_j$ to obtain
\begin{subequations}
\begin{align}
    &\dot{\bq}_b = \bA_b^{-1} \biggl(\bh_{com} - \bA_j \dot{\bq}_j \biggr) \\
    &\dot{\bq}_j = \bv_j
\end{align} \label{eq:dynamics}
\end{subequations}
Note that the dynamics presented in \eqref{eq:dynamics} is nonlinear and under-determined (since $\bA(\bq)$ is a fat matrix). The system has multiple feasible solutions for a given input $\bu$. 

In this paper, we adopt the nonlinear MPC proposed in \cite{sleiman2021unified}. In general, the nonlinear MPC problem with a horizon $N$ can be formulated as
\begin{align}
    \min_{\bx,\bu} \hspace{2mm} &\sum_{k=0}^{N-1} \bigl\{||\bx_{k+1} - \bx_{k+1,ref}||_{\bQ_k} + ||\bu_k||_{\bK_k} \bigr\} \nonumber \\
    \textrm{s.t.}  \nonumber \\
    & \dot \bx = \bff_q(\bx, \bu, t) \qquad \qquad \qquad \nonumber \\ 
    &\mathbf{C}_{c_i}\bu_{c_i,k} \leq \mathbf{0} \nonumber \\
    & \bD_{c_i}\bu = \mathbf{0}\nonumber \\
    &\bu \in \bU \nonumber
    \label{eq:NMPC}
\end{align}

where $\bx_k$ and $\bu_k$ are the state and control input respectively at time step $k$, $\bQ_k$ and $\bK_k$ are diagonal positive definite weight matrices, $\bff_q$ is the hybrid centroidal model for the quadruped, $\mathbf{C}_{c_i}$ is the linear friction matrix for the corresponding reaction force $\bu_{c_i,k}$ in contact, and $\bD_{c_i}$ is selection matrix on the reaction forces in swing phase. The optimization is solved in closed-loop in a receding horizon fashion using the Sequential Linear Quadratic (SLQ) technique \cite{diehl2005real}.

%%%%%%%%%%%%%%% low level control %%%%%%%%%%%%%%%%%%%%%%%%%%%%%%%
\subsection{Low-level Controller}
The low-level controller is designed to track the optimal reference plans. This is formulated through a hierarchical quadratic program (QP) which optimizes the actuator torques under a higher fidelity model, friction, and actuator constraints. To map the MPC output to the QP, extra information is extrapolated such as the swing feet trajectories, which can be obtained through a simple kinematic relation of task to joint space acceleration
\begin{equation}
    \bJ_{c_i}\Ddot{\bq}_j + \Dot{\bJ}_{c_i}\Dot{\bq}_j = \Ddot{\br}_{c_i}
\end{equation}
where $\bJ_{c_i}$ is the swing leg Jacobian and $\Ddot{\br}$ is the end-effector acceleration.
The optimized trajectory from the QP is tracked using a drive controller where the torque actuated ($\btau_a$) is defined by the following
\begin{equation}
    \btau_a = \btau_j + \bK_p(\bq^\star - \bq_j) + \bK_d(\dot{\bq}^\star - \dot{\bq}_j)
\end{equation}
where $\btau_j$ is the expected joint torque, $\dot{\bq}^\star$ and $\bq^\star$ are the optimized joint reference states,  
More details on the QP formulation can be referenced to \cite{sleiman2021unified}.

\subsection{Simulation Results}
\begin{figure*}
    \centering
  \includegraphics[width=1\textwidth]{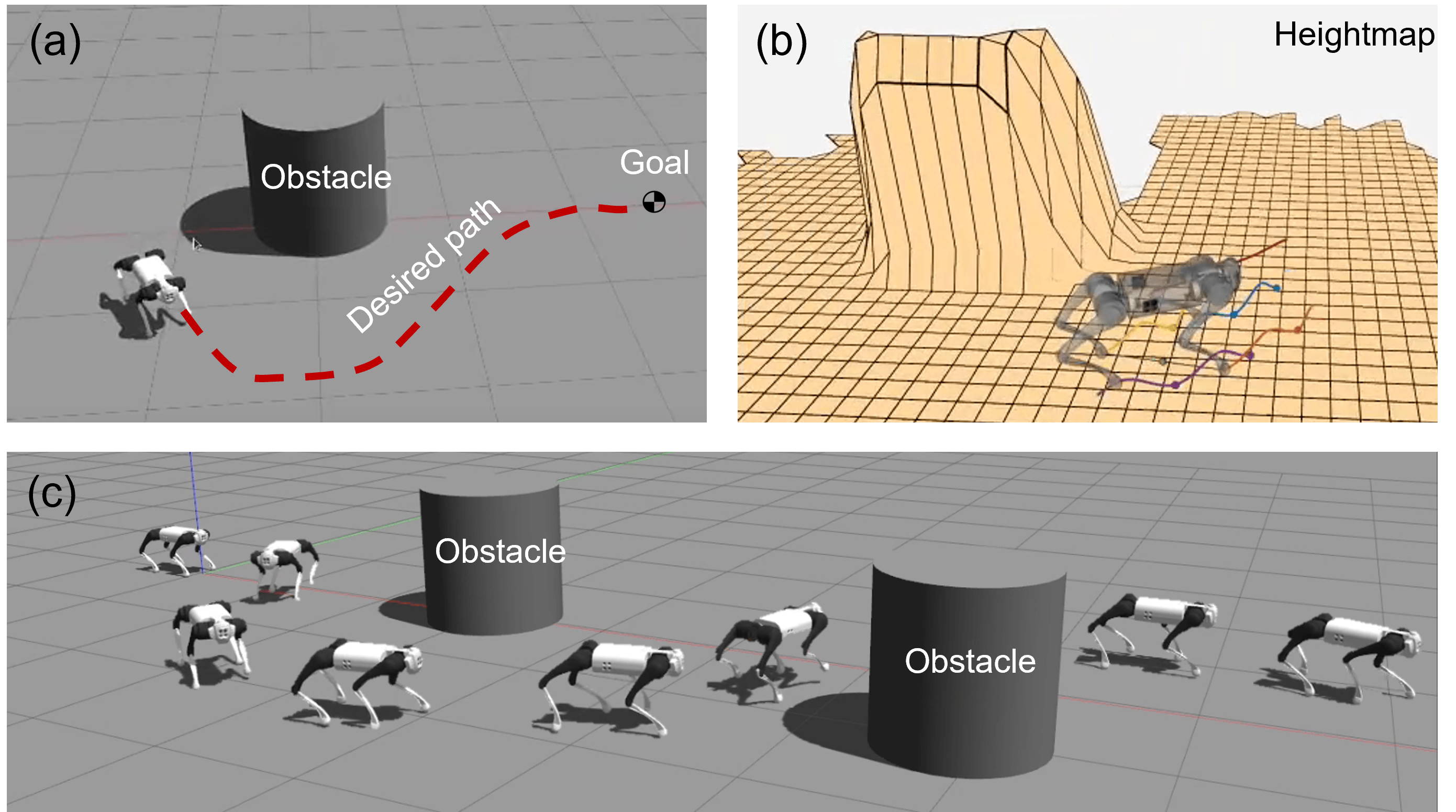}
  \caption{Quadruped robot executing the proposed global plan. (a) Robot executing a safe motion plan in an environment with a single obstacle and (b) the corresponding height map representation. (c) safe motion plan for an environment with two obstacles.}
  \label{fig:quad_results}
\end{figure*}
Simulation results are shown using Gazebo with RVIZ as the visual interface, as shown in Figure \ref{fig:quad_results}. In the following experiment, we define a target set $\bX_T = (10,0)$. In Figure  \ref{fig:quad_results}a, we set up an environment with a cylindrical obstacle $\bX_{u}$ centered at (5,0) with a radius of {\color{black}$r = 0.5$ (and enlarged appropriately to account for the size of the robot)}, while in Figure \ref{fig:quad_results}c, we use two cylindrical obstacles centered at (3,0.1) and (7,-1) respectively. Using these parameters, a density function is constructed. The reference trajectory is obtained as a feedback control using \eqref{eq:control}. By forward integrating the feedback controller for a fixed horizon, our feedback controller is fed into the NMPC as a reference trajectory. Note that for an increase in performance, we use a first-order filter and a moving average filter to smoothen the trajectories generated by the density planner.

\begin{figure}
    \centering
    \includegraphics[width = 0.6\linewidth]{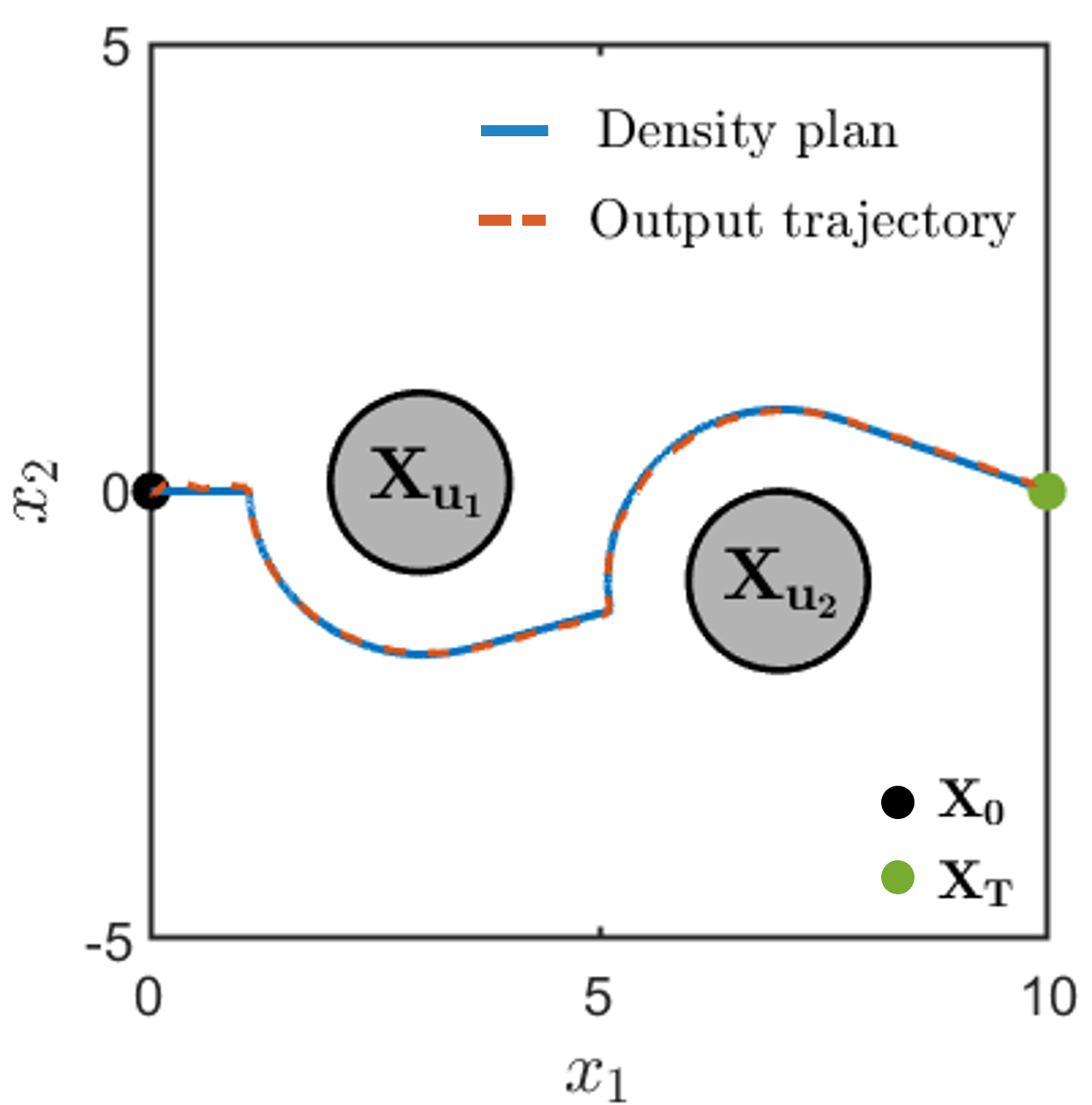}
    \caption{Quadruped robot state trajectory (output) with desired density-based plan.}
    \label{fig:quad_simulation}
\end{figure}

The resulting trajectory of the center of mass of the quadruped in comparison with the center of mass trajectory from the density is shown in Figure \ref{fig:quad_simulation}. Despite the model differences, we see that the model predictive controller tracks the reference trajectory well.

\begin{figure}
    \centering
    \includegraphics[width = \linewidth]{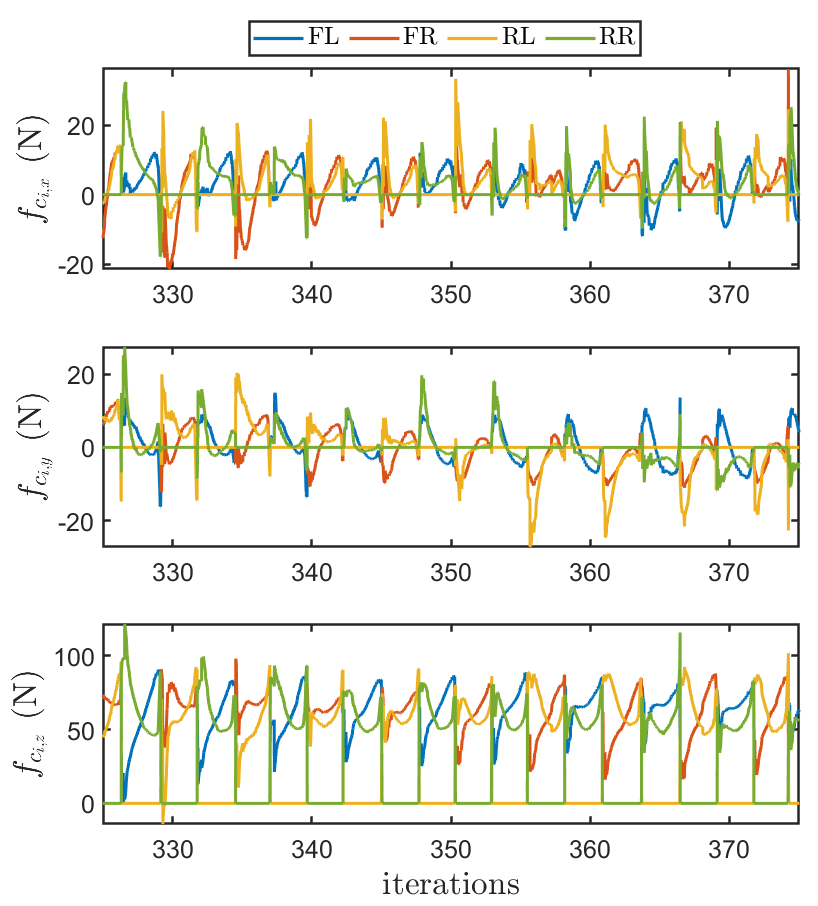}
    \caption{(FL: front left, FR: front right, RL: rear left, RR: rear right) GRFs produced by the feet. Note the y-direction reaction force ($F_y$) changes the direction track density-based plan.}
    \label{fig:quad_simulation_grf}
\end{figure}

Correspondingly, we are able to see the resulting optimized reaction force $\bu_k$ to track the reference trajectory from the density plan in Figure \ref{fig:quad_simulation_grf}. Note, when the quadruped tracks the reference trajectory in Figure \ref{fig:quad_simulation}, the optimized ground reaction force $\bu$ is seen to be nonzero in the transverse direction. This is due to ground reaction force being the main component in propelling the quadruped in the transverse plane, hence nonzero forces for $\bff_{c_i,x}$ and $\bff_{c_i,y}$. Moreso, for the quadruped traverse through the trajectory of the density plan with nonzero curvature, the optimized ground reaction force must invert signs. This is highlighted during $n \in [330, 350]$ iterations, where the quadruped is traversing around the obstacle.

\subsection{Hardware Implementation}
The proposed density-based planner is implemented in hardware on a Unitree GO1 robot. The robot was initialized in a lab environment with full knowledge of the environment. A safe trajectory was generated using the global planner and density functions which was successfully tracked by the NMPC and low-level controller. A video of the implementation is available on YouTube\footnote{\href{https://www.youtube.com/watch?v=gJH6RTcHrfg}{https://www.youtube.com/watch?v=gJH6RTcHrfg}}.

%%%%%%%%%%%%%%%%%%%%%%%%%%%%%%%%%%%%%%%%%%%%%%%%%%%%%%%%%%%%%%%%%%%%%%%%%%%%%%%% 
% \section{Limitations} \label{sec:limitations}
% Some limitations of the current framework are listed below
% \begin{itemize}
%     \item The density-based planner works only for binary environments. There is no distinction for the degree of traversability and hence it cannot be easily extended to off-road navigation. 
%     \item This formulation only works for holonomic systems. Although, there are several ways to extend for non-holonomic cases.
%     \item The formulation relies on global information for the environment and hence cannot incorporate local sensing information to form density functions online.
%     \item The quality of trajectories obtained from the global planner can be improved by passing it through a trajectory optimizer.
% \end{itemize}

%%%%%%%%%%%%%%%%%%%%%%%%%%%%%%%%%%%%%%%%%%%%%%%%%%%%%%%%%%%%%%%%%%%%%%%%%
\section{CONCLUSIONS}\label{sec:conclusions}
In this work, we develop a safe motion planning architecture for quadruped locomotion. We use density functions to design a safe reference trajectory for the robot. The trajectories are obtained as a positive gradient of the proposed density function, giving a closed loop feedback form. The proposed algorithm is integrated with a nonlinear MPC and low-level controller from the OCS2 framework. Simulation results show that the robot is able to track safe reference trajectories provided by the density-based motion planning framework. 

Some of the current limitations of this work is that the density formulation is limited to a binary representation of the environment, safe and unsafe. Therefore, there is no distinction for the degree of traversability, limiting the extension in off-road environments. Additionally, the current formulation assumes a holonomic system, hence, limiting the class of systems for this feedback controller. Lastly, the quality of trajectories can be improved, and therefore a filter to smoothen state and control trajectory will greatly increase the performance of the planner. Future works will look to integrate a trajectory optimizer to smoothen the trajectory and extend this framework to non-holonomic systems and off-road navigation.

\addtolength{\textheight}{-12cm}   % This command serves to balance the column lengths

%Bibliography
\bibliographystyle{unsrt}  
\bibliography{references}  
\end{document}